\documentclass{article}
\usepackage{spconf,amsmath,graphicx,hyperref}
\usepackage{booktabs}
\usepackage{tabularx}
\usepackage{caption}
\usepackage{array} 
\usepackage{colortbl}
\usepackage{xcolor}
\usepackage{makecell} 
\usepackage{amsmath,amssymb}
\usepackage{algpseudocode}
\usepackage{algorithm}
\usepackage[T1]{fontenc}
\usepackage[utf8]{inputenc}
\usepackage{tikz}
\usetikzlibrary{positioning, arrows.meta, shapes.geometric, calc}

\newcolumntype{Y}{>{\centering\arraybackslash}X}
\definecolor{bestcolor}{RGB}{255,200,120}
\DeclareUnicodeCharacter{03B3}{\ensuremath{\gamma}}

\DeclareUnicodeCharacter{0097}{---} % 把 U+0097 映射成 LaTeX 的长破折号

\title{Large Language Models as Discounted Bayesian Filters}
\name{Jensen Zhang \qquad Jing Yang   \qquad Keze Wang}
\address{Sun Yat-sen University}

\begin{document}
\maketitle
\begin{abstract}
Large Language Models (LLMs) demonstrate strong few-shot generalization through in-context learning (ICL), yet their reasoning in dynamic and stochastic environments remains opaque. Prior studies mainly address static tasks, overlooking the online adaptation required when beliefs must be continuously updated—a key capability for LLMs as world models or agents. We introduce a Bayesian filtering framework to evaluate online inference in LLMs. Our probabilistic probe suite spans multivariate discrete (e.g., dice rolls) and continuous (e.g., Gaussian) distributions, where ground-truth parameters shift over time. 
We find that while LLMs’ belief updates resemble Bayesian posteriors, they are more accurately described by an exponential forgetting filter with a model-specific discount factor $\gamma<1$. This reveals systematic discounting of older evidence, varying significantly across architectures. Although inherent priors are often miscalibrated, the updating mechanism itself is structured and principled. We validate these findings in a simulated agent task and present prompting strategies that effectively recalibrate priors with minimal cost. 
\end{abstract}


\section{Introduction}
\label{sec:intro}
In-context learning (ICL) represents a remarkable capability of large language models (LLMs)\cite{z1,z2,z3,z4}, enabling rapid adaptation to novel tasks based solely on a handful of examples provided in their prompts, without explicit gradient updates. While the empirical success of ICL underpins modern prompting techniques, its fundamental mechanism remains largely opaque. A key open question is whether ICL constitutes structured statistical reasoning analogous to Bayesian inference or is merely sophisticated pattern recognition.

A promising perspective frames ICL as implicit Bayesian inference, where models iteratively update latent belief states based on contextual evidence\cite{z5,z6,zhang2025kabb}. However, existing foundational studies predominantly investigate \emph{static} environments, assuming stationary data-generating distributions. Such an assumption neglects a crucial aspect of real-world intelligence: the necessity to operate effectively within \emph{non-stationary} environments. In these dynamic contexts, agents must continuously integrate new information while systematically discounting—or “forgetting’’—outdated evidence whose relevance diminishes over time. This capacity for online adaptation is critical for deploying LLMs as reliable world models or autonomous agents, yet remains under-explored.
\begin{figure}[tb]
 \centering
 \includegraphics[width=0.95\columnwidth]{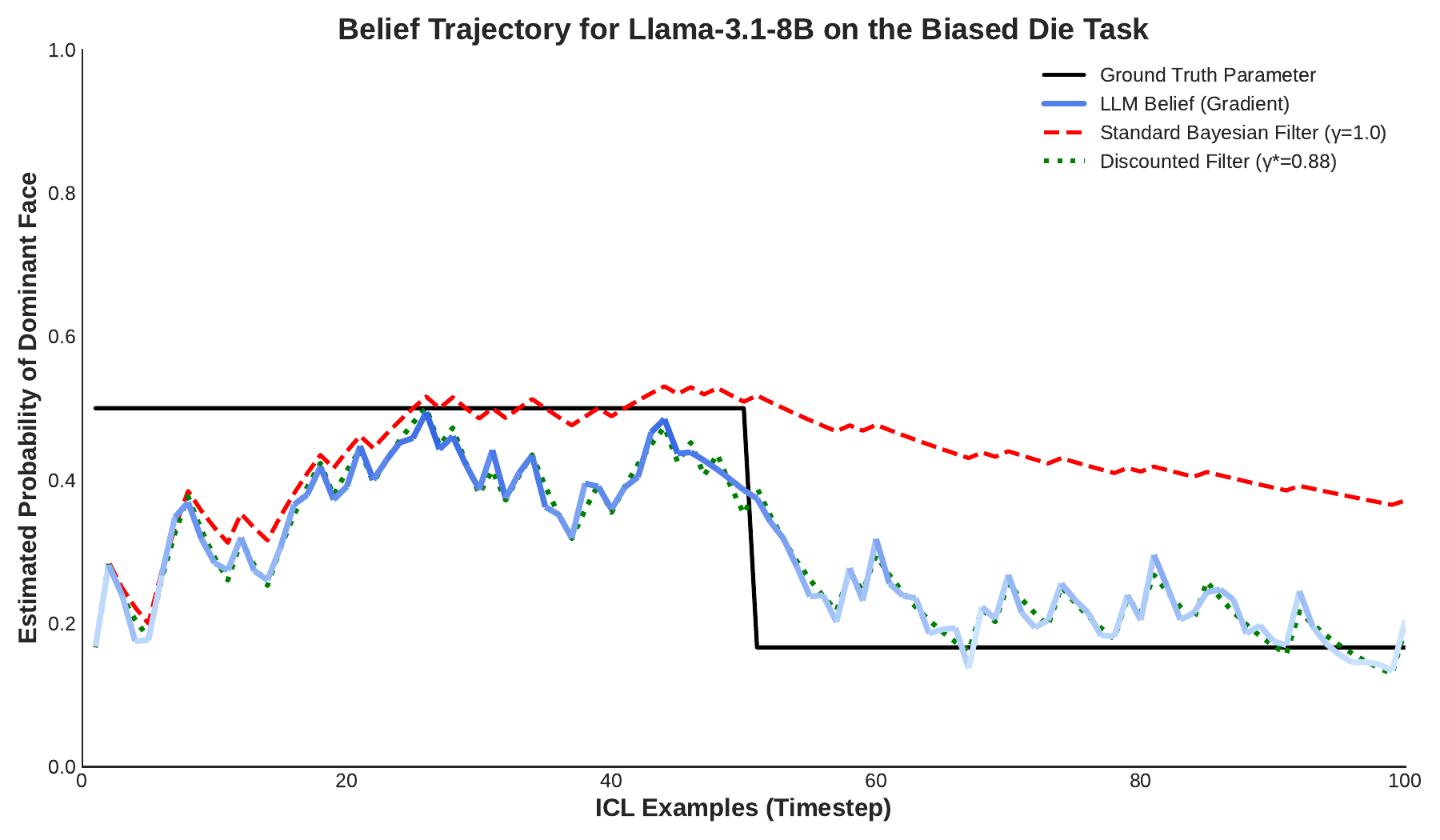}
\caption{Belief trajectory on the Biased Die task: LLM estimates (solid) vs. discounted filters (dashed), with $\gamma^*$ giving the closest match.}
 \label{fig:trajectory1}
\end{figure}
In this work, we propose a new theoretical perspective: \textbf{we conceptualize ICL in Transformers as online Bayesian filtering characterized by systematic discounting of past evidence}. Central to our thesis is the assertion that LLMs, when confronted with sequences of evolving evidence, do not function as ideal Bayesian observers. Instead, we hypothesize that their behavior incorporates an intrinsic forgetting mechanism, likely arising from architectural elements of Transformers\cite{xie2022explanation,z7}. To formalize and empirically evaluate this hypothesis, we introduce an analytical framework built around fitting a \textbf{discount factor $\gamma \in (0,1]$} that minimizes the Kullback-Leibler (KL) divergence between a model's predictive distribution and a theoretical Bayesian filter's. We investigate this dynamic behaviour through a controlled \textbf{probabilistic probe suite}\cite{xie2022explanation,chen2003bayesian,chen2003signal}, involving tasks such as biased die rolling and Gaussian mean estimation, where the ground-truth parameters undergo sudden shifts, compelling models to adapt their posterior beliefs.

Through our integrated experimental and analytical approach, we present the following contributions:
\begin{enumerate}
  \item We establish a novel framework for interpreting ICL as online Bayesian filtering. Our analysis provides the first quantitative demonstration that LLMs systematically discount historical information, a behaviour accurately modelled by fitting a stable, model-specific discount factor~$\gamma$ via a principled optimization process.
  
  \item The proposed framework enables a nuanced decomposition of predictive errors into what we term \textbf{Update Divergence} and \textbf{Model Misspecification Divergence}. We show that observed performance limitations are predominantly attributable to the latter, which stems from miscalibrated priors and intrinsic discounting, rather than a flawed updating process itself.
  
  \item We further investigate the internal mechanisms underlying this discounting behaviour, revealing through \textbf{correlation analysis} that inferential quality is decoupled from the raw magnitude of attention scores. This finding points to a complex architectural basis for evidential forgetting that extends beyond straightforward attention allocation.
\end{enumerate}

\section{Related Work}
\label{sec:relatedwork}
In-context learning (ICL) in large language models (LLMs) \cite{chan2022data}has been extensively studied as a form of implicit statistical inference, particularly through a Bayesian lens. Foundational works, such as those by Xie et al.~\cite{xie2022explanation} and Aky"{u}rek et al.~\cite{akyurek2022what}, frame ICL as approximating Bayesian posterior updates or gradient descent on latent functions, enabling few-shot generalization in stationary environments. However, these perspectives assume fixed data distributions, overlooking the challenges of non-stationary settings where beliefs must adapt online to shifting parameters---a gap our discounted filtering framework addresses by introducing systematic evidence forgetting.

Recent efforts have begun exploring dynamic adaptation in LLMs, including continual learning approaches~\cite{Kirkpatrick_2017} that mitigate catastrophic forgetting via architectural modifications or replay buffers. For instance, methods like self-synthesized rehearsal~\cite{guo2024mitigating} generate synthetic data to preserve knowledge across tasks. Yet, these often focus on task-specific retention rather than principled probabilistic updating under uncertainty.Our work diverges by providing a unified Bayesian filtering model with a fitted discount factor $\gamma<1$, quantifying deviations from ideal inference and linking them to Transformer attention mechanisms, thus bridging static ICL theory with online reasoning. 

\section{Methodology}
\label{sec:methodology}

Our methodology is designed to empirically test if LLM in-context learning emulates online Bayesian inference with evidence discounting. We first define our theoretical model and the non-stationary tasks used for probing, then detail the quantitative analysis, summarized in Algorithm~\ref{alg:main}.

\subsection{Framework and Experimental Probes}
\textbf{Discounted Bayesian Filtering.} We model the "forgetting" of past evidence using a discounted Bayesian filtering framework\cite{west1997dlm}. This introduces a discount factor, $\gamma \in (0, 1]$, which tempers the posterior belief $p_{t-1}(\theta | D_{1:t-1})$ before it is updated with new evidence $D_t$:
\begin{equation}
    p'_{t-1}(\theta) \propto [p_{t-1}(\theta | D_{1:t-1})]^{\gamma}
    \label{eq:discount}
\end{equation}
Here, $\gamma=1$ represents a standard Bayesian filter with perfect memory, while $\gamma \to 0$ approaches a memoryless state.

\noindent\textbf{Probabilistic Probe Suite.} To test online adaptation, we designed two non-stationary tasks over a sequence of $T=100$ observations, with a parameter changepoint at $t=51$:
\textbf{Biased Die (Discrete):} A 6-sided die where the dominant face abruptly changes from one to another. \textbf{Gaussian Mean (Continuous):} Samples from $\mathcal{N}(\mu, 1)$, where the mean $\mu$ shifts from $2.0$ to $-2.0$.
At each timestep $t$, we elicit the LLM's predictive distribution $\hat{\mathbf{p}}_{\text{LLM}, t}$ by providing the history $D_{1:t-1}$ and applying softmax to the output logits for all possible next outcomes.

\subsection{Quantitative Analysis}
Our analysis quantitatively connects the LLM's behavior to our framework. The core procedure is outlined in Algorithm~\ref{alg:main}.
\noindent\textbf{1. Quantifying Discounting ($\gamma^*$):} We fit an optimal discount factor $\gamma^*$ by minimizing the total KL divergence between the LLM's predictive sequence and that of the discounted filter:
\begin{equation}
    \gamma^* = \arg\min_{\gamma \in (0, 1]} \sum_{t=1}^{T} \mathcal{D}_{\text{KL}}\left( \hat{\mathbf{p}}_{\text{LLM}, t} \parallel \mathbf{p}_{\text{Bayes}, t}(\gamma) \right)
    \label{eq:fitting}
\end{equation}
We solve this using the L-BFGS-B algorithm. A fitted $\gamma^* < 1$ provides direct evidence of discounting.

\noindent\textbf{2. Decomposing Predictive Error:} We diagnose the source of errors by decomposing the total divergence into two components:
\textbf{Update Divergence ($\mathcal{D}_{\text{Update}}$):} $\mathcal{D}_{\text{KL}}(\hat{\mathbf{p}}_{\text{LLM}} \parallel \mathbf{p}_{\text{Bayes}}(\gamma^*))$. Measures if the LLM is a principled discounted updater.
\textbf{Model Misspecification Divergence ($\mathcal{D}_{\text{ModelSpec}}$):} $\mathcal{D}_{\text{KL}}(\mathbf{p}_{\text{Bayes}}(\gamma^*) \parallel \mathbf{p}_{\text{Truth}})$. Measures the inherent error of the best-fit model.

\noindent\textbf{3. Analyzing Architectural Basis:} We investigate the role of attention by computing the Pearson correlation $\rho$ between the aggregate final-layer attention score on past evidence ($A_t$) and the step-wise Update Divergence ($E_t$), to find a potential architectural basis for the discounting behavior.

\begin{algorithm}[t]
\caption{Fitting $\gamma^*$ and Decomposing Error}
\label{alg:main}
\begin{algorithmic}[1]
\State \textbf{Input:} LLM $\mathcal{M}$, Task observations $D_{1:T}$
\State \textbf{Output:} $\gamma^*, \mathcal{D}_{\text{Update}}, \mathcal{D}_{\text{ModelSpec}}$
\State
\State \Comment{1. Elicit LLM's belief trajectory}
\State $\hat{\mathbf{P}}_{\text{LLM}} \gets \{\}$
\For{$t=1$ to $T$}
    \State $\hat{\mathbf{p}}_{\text{LLM}, t} \gets \text{Softmax}(\mathcal{M}(D_{1:t-1}))$
    \State Append $\hat{\mathbf{p}}_{\text{LLM}, t}$ to $\hat{\mathbf{P}}_{\text{LLM}}$
\EndFor
\State
\State \Comment{2. Find optimal discount factor via optimization}
\State Define objective $L(\gamma) = \sum_{t=1}^{T} \mathcal{D}_{\text{KL}}(\hat{\mathbf{p}}_{\text{LLM}, t} \parallel \mathbf{p}_{\text{Bayes}, t}(\gamma))$
\State $\gamma^* \gets \operatorname*{argmin}_{\gamma \in (0, 1]} L(\gamma)$ \Comment{Using L-BFGS-B}
\State
\State \Comment{3. Decompose error with the optimal $\gamma^*$}
\State $\mathcal{D}_{\text{Update}} \gets \frac{1}{T} L(\gamma^*)$
\State $\mathcal{D}_{\text{ModelSpec}} \gets \frac{1}{T} \sum_{t=1}^{T} \mathcal{D}_{\text{KL}}(\mathbf{p}_{\text{Bayes}, t}(\gamma^*) \parallel \mathbf{p}_{\text{Truth}, t})$
\State \textbf{return} $\gamma^*, \mathcal{D}_{\text{Update}}, \mathcal{D}_{\text{ModelSpec}}$
\end{algorithmic}
\end{algorithm}

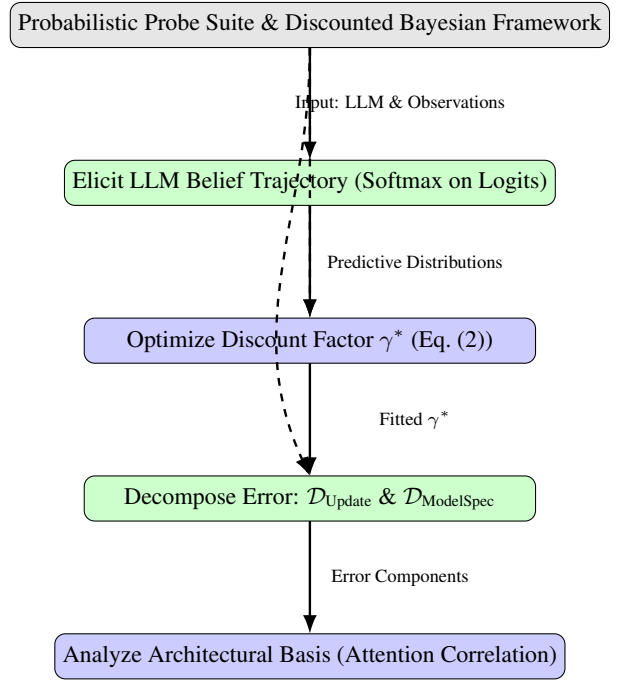
\begin{figure}[t]
\centering
\begin{tikzpicture}[
  node distance=15mm,
  every node/.style={font=\small},
  >=Latex,
  box/.style={draw, rounded corners, align=center, inner sep=3pt, outer sep=0pt,
              minimum width=6.0cm, minimum height=6mm},
  boxJ/.style={box, fill=gray!20},
  boxG/.style={box, fill=green!20},
  boxB/.style={box, fill=blue!20},
  edge/.style={->, line width=0.9pt},
  edgeDashed/.style={edge, dashed},
  lab/.style={font=\scriptsize, inner sep=1pt}
]
  % ==== 节点 ====
  \node[boxJ] (top) {Probabilistic Probe Suite \& Discounted Bayesian Framework};
  \node[boxG, below=of top] (elicit) {Elicit LLM Belief Trajectory (Softmax on Logits)};
  \node[boxB, below=of elicit] (optimize) {Optimize Discount Factor $\gamma^*$ (Eq.~\eqref{eq:fitting})};
  \node[boxG, below=of optimize] (decompose) {Decompose Error: $\mathcal{D}_{\text{Update}}$ \& $\mathcal{D}_{\text{ModelSpec}}$};
  \node[boxB, below=of decompose] (analyze) {Analyze Architectural Basis (Attention Correlation)};

  % ==== 主链 ====
  \draw[edge] (top) -- (elicit)   node[lab, midway, xshift=12mm] {Input: LLM \& Observations};
  \draw[edge] (elicit) -- (optimize) node[lab, midway, xshift=14mm] {Predictive Distributions};
  \draw[edge] (optimize) -- (decompose) node[lab, midway, xshift=14mm] {Fitted $\gamma^*$};
  \draw[edge] (decompose) -- (analyze) node[lab, midway, xshift=12mm] {Error Components};

  % ==== 回路（可选，如果有循环，可添加；这里假设线性流程，无回路） ====
  % \draw[edge, bend left=25] (analyze.east) to[out=0, in=-10] (optimize.south east);

  % ==== 虚线（类似示例中的JAX加速，这里可模拟为优化加速） ====
  \draw[edgeDashed] (top.south) to[out=-90, in=90]  (optimize.north);
  \draw[edgeDashed] (top.south) to[out=-90, in=115] (decompose.north);
\end{tikzpicture}
\caption{Overview of the proposed methodology for evaluating LLM online inference behavior, illustrating the unified process of belief elicitation, discount factor optimization, error decomposition, and architectural analysis.}
\label{fig:methodology_pipeline}
\end{figure}
\section{Experiments}
\label{sec:experiment}

We empirically validate our hypothesis using the setup from Section~\ref{sec:methodology}. We evaluated the base and instruction-tuned variants of Llama-3.1-8B\cite{grattafiori2024llama3herdmodels}, Mistral-7B\cite{jiang2023mistral7b}, and Gemma-2-2B\cite{gemmateam2024gemma2improvingopen} on the \textbf{Biased Die\cite{murphy2012ml}} and \textbf{Gaussian Mean\cite{bishop2006prml}} probes, fitting an optimal discount factor $\gamma^*$ and decomposing predictive error for each model-task pair.

\begin{figure}[t]
 \centering
 \includegraphics[width=0.52\textwidth]{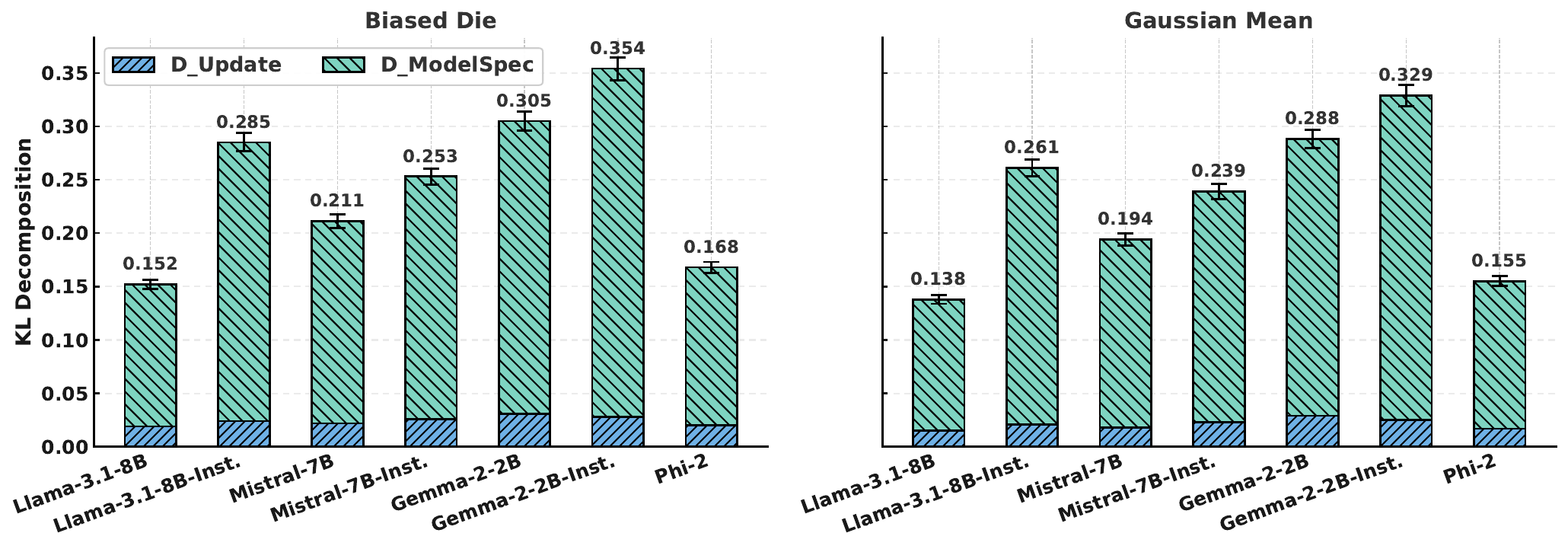} % Placeholder name
 \caption{Decomposition of Total KL Divergence. Low $\mathcal{D}_{Update}$ values indicate that LLMs closely follow a discounted filtering process, whereas high $\mathcal{D}_{ModelSpec}$ highlights the inherent limitations of this simplified approximation.}
 \label{fig:error_decomposition}
\end{figure}

\subsection{Baselines and Theoretical Context}
Our work builds on, yet diverges from, prior perspectives on ICL\cite{adams2007bayesianonlinechangepointdetection}. While foundational views frame ICL as implicit Bayesian inference under stationary assumptions (equivalent to $\gamma=1$), others identify systematic deviations from this ideal without proposing an alternative mechanism~(Figure~\ref{fig:trajectory1}). Our framework offers a specific mechanistic account for these deviations, testing whether the intrinsic belief updating of a \textit{single} LLM is captured by a simple, principled discounting process. This provides a more fundamental view of adaptation than active filtering or ensemble gating approaches.

\subsection{Results and Analysis}
\begin{figure}[tb]
 \centering
 \includegraphics[width=0.95\columnwidth]{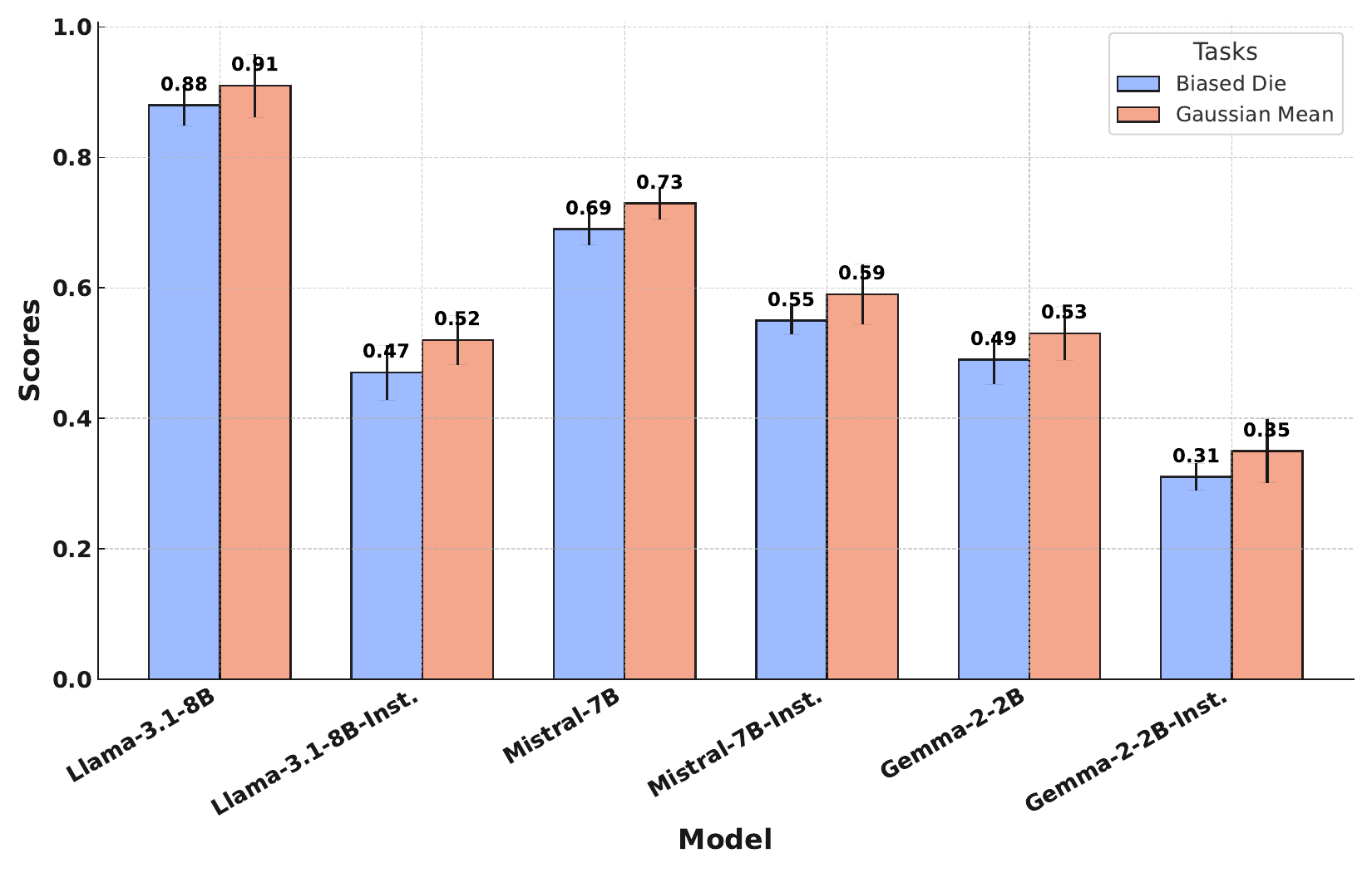} % Placeholder name
 \caption{Optimal Discount Factors $\gamma^*$ by Model and Task. Values consistently below 1 indicate systematic discounting of past evidence.}
 \label{fig:gamma_results}
\end{figure}

\textbf{Fitted Discount Factors ($\gamma^*$): Evidence for Forgetting.}
The optimal discount factors (Figure~\ref{fig:gamma_results}) provide strong evidence against the classical Bayesian model. For all models, the fitted $\gamma^*$ is significantly below 1, contradicting the perfect-memory assumption and indicating that LLMs inherently discount past evidence. We observe consistent patterns: instruction-tuned models exhibit lower $\gamma^*$ (stronger discounting), and each model family shows a characteristic discounting rate across tasks.

\noindent\textbf{Error Decomposition: Pinpointing the Source of Deviations.}
The error decomposition in Figure~\ref{fig:error_decomposition} reveals \textit{why} the discounted model is superior. The majority of the total error ($\mathcal{D}_{Total}$) is from \textbf{Model Misspecification Divergence} ($\mathcal{D}_{ModelSpec}$), which reflects the inherent limits of any simplified model. Crucially, the \textbf{Update Divergence} ($\mathcal{D}_{Update}$)—the deviation from their own best-fit model—is consistently small (typically $<13\%$ of total error).
This result is key: it suggests LLMs are not flawed Bayesian updaters but proficient \textbf{discounted updaters}. Their behavior is principled, and the failure of the standard Bayesian model ($\gamma=1$) is its inability to forget outdated evidence after a changepoint—a limitation the fitted $\gamma^*$ overcomes.
\subsection{Architectural Basis: Attention as a Modulator of Inferential Quality}
\begin{figure}[h!]
 \centering
 \includegraphics[width=0.99\columnwidth]{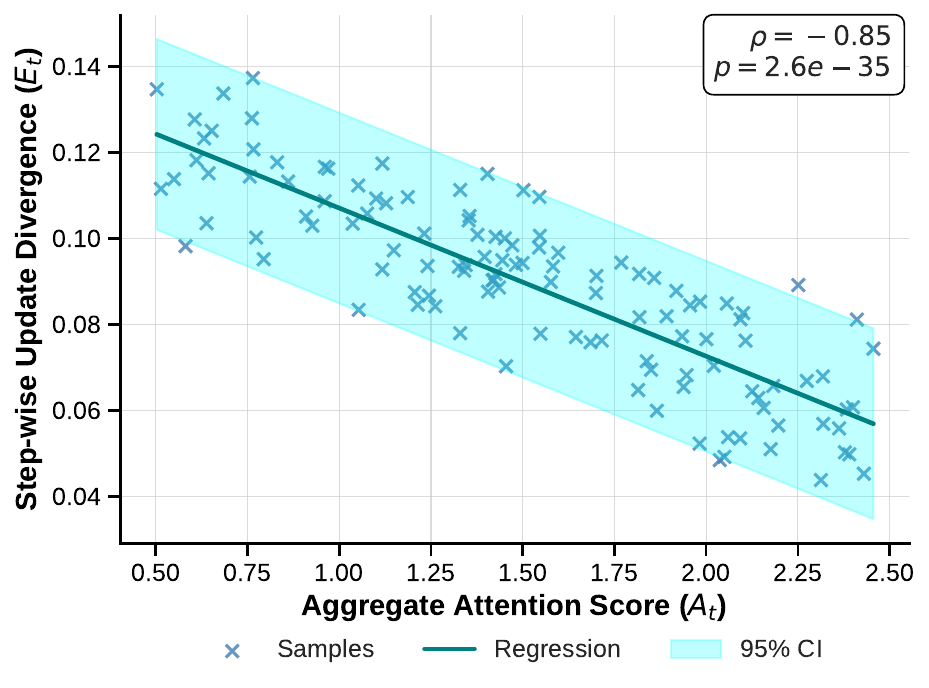} 
\caption{Attention vs. Update Divergence ($\rho = -0.85$) for Llama-3.1-8B on Biased Die.}
 \label{fig:attention_scatter}
\end{figure}

Our final analysis investigates the architectural basis for this behavior by examining the link between the Transformer's attention\cite{vaswani2023attentionneed} and inferential quality\cite{olsson2022incontextlearninginductionheads}. We test the hypothesis that attention allocated to historical context correlates with the stability of the belief update.
For Llama-3.1-8B on the Biased Die task, we plotted the step-wise Update Divergence ($E_t$) against the Aggregate Attention Score on past evidence ($A_t$), as shown in Figure~\ref{fig:attention_scatter}. The results reveal a strong and highly significant negative correlation ($\rho = -0.85, p \approx 2.6 \times 10^{-35}$). This finding indicates that greater aggregate attention to past evidence is directly associated with more principled and stable belief updates (lower divergence). The attention mechanism thus acts as a primary modulator of inferential fidelity, providing a compelling architectural explanation for the observed belief-updating behavior.
\subsubsection{Internal Dynamics: Clustering of Hidden State Representations}
\begin{figure}[h!]
 \centering
 \includegraphics[width=\columnwidth]{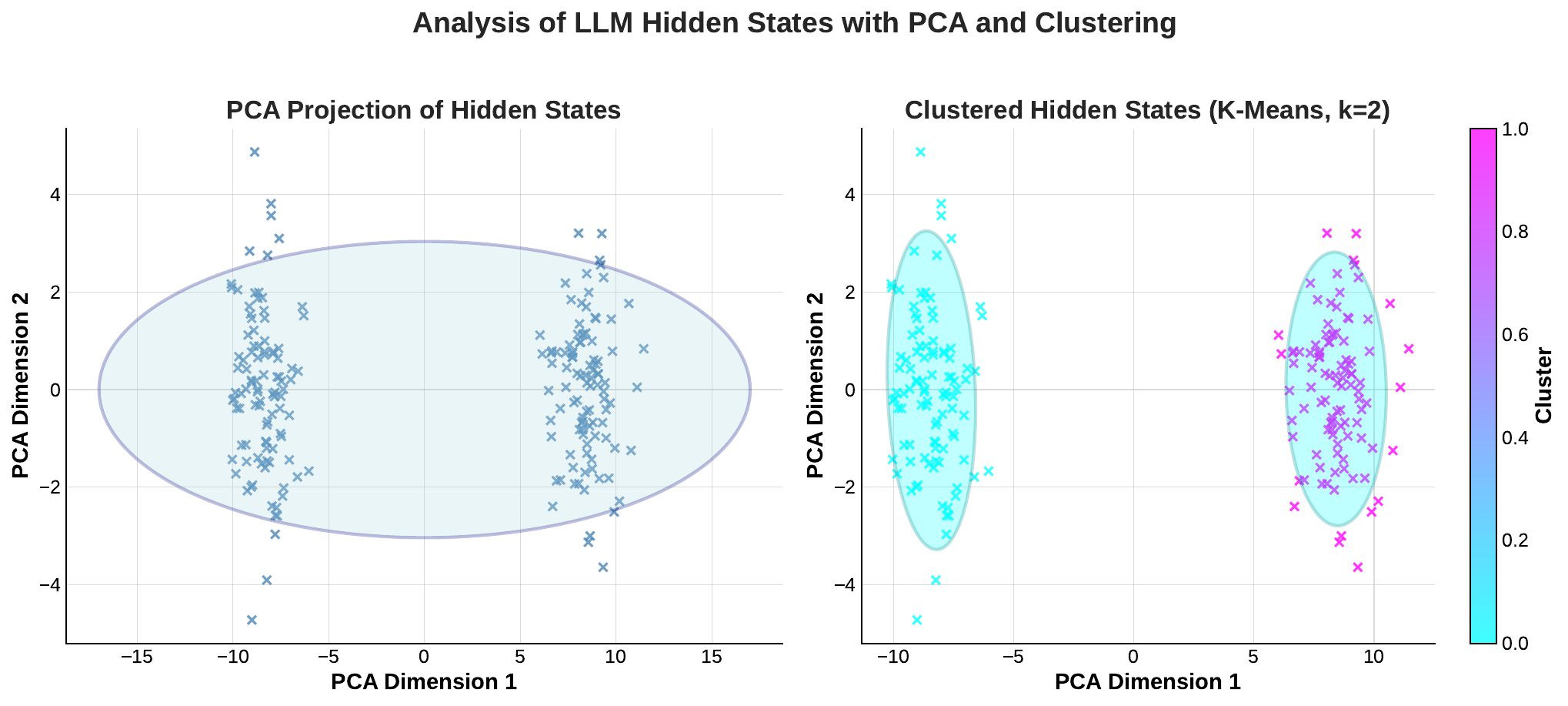} 
 \caption{PCA visualization of final-layer hidden states across 100 timesteps. \textbf{(Left)} The 2D projection reveals two clusters. \textbf{(Right)} K-Means clustering ($k=2$) rediscovers the changepoint, perfectly separating the two phases.}
 \label{fig:hidden_state_pca}
\end{figure}

To directly visualize the model's internal adaptation, we examined the trajectory of its hidden state representations. If online adaptation occurs, these states should cluster according to the environment's phases. 
For a representative model, we extracted the final-layer hidden states across $T=100$ timesteps and projected them into 2D via Principal Component Analysis (PCA). 
We then applied K-Means clustering ($k=2$) to the original high-dimensional vectors. The resulting partition (Figure~\ref{fig:hidden_state_pca}, right) aligns almost perfectly with the pre- ($t \le 50$) and post-changepoint ($t > 50$) phases, despite the algorithm having no prior task knowledge. This demonstrates that the model’s hidden states dynamically track distributional shifts, providing representational evidence of the online belief updating central to our thesis.
\section{Conclusion}
\label{sec:conclusion}
In this work, we introduced a discounted Bayesian filtering perspective to analyze the online inference behavior of large language models. Through a comprehensive suite of dynamic probabilistic probes, we showed that LLMs consistently discount past evidence when updating beliefs, a behavior quantified by a fitted discount factor $\gamma < 1$. Our error decomposition further revealed that predictive errors mainly stem from model misspecification rather than faulty updating, underscoring that LLMs function as principled discounted updaters. This characterization clarifies when and why LLM predictions drift under non-stationarity, and offers a principled basis for diagnosing failures and designing adaptive correction or calibration mechanisms. These findings provide both a robust theoretical lens and a practical toolkit for understanding, evaluating, and potentially improving the reasoning of LLMs in uncertain and non-stationary environments.

% -------------------------------------------------------------------------
\bibliographystyle{IEEEbib}
\bibliography{strings,refs}

\end{document}